\documentclass[12pt]{article}

\usepackage{natbib}
\usepackage{url} 
\usepackage{amsmath}
\usepackage{amssymb}
\usepackage{amsfonts}
\usepackage{multirow}
\usepackage{amsthm}
\usepackage{bm}
\usepackage{subcaption}
\usepackage{diagbox,color}
\usepackage{float}
\usepackage{graphicx}
\usepackage[title]{appendix}
\usepackage{setspace}
\usepackage{comment}

\newcommand{\blind}{0}

\DeclareMathOperator*{\argmin}{arg\,min}

\newtheorem{theorem}{Theorem}

\theoremstyle{definition}
\newtheorem{definition}{Definition}

\newtheorem{assumption}{Assumption}
\let\oldassumption\assumption
\renewcommand{\assumption}{\oldassumption\normalfont}

\begin{document}

\def\spacingset#1{\renewcommand{\baselinestretch}%
{#1}\small\normalsize} \spacingset{1}


\if0\blind
{
  \title{\bf Functional Nonlinear Learning}

  \author{Haixu Wang\hspace{.2cm}
    and Jiguo Cao \\
    Department of Statistics and Acturial Science\\ 
    Simon Fraser University, Burnaby, BC, Canada}
\date{}
  \maketitle
} \fi

\if1\blind
{
  \bigskip
  \bigskip
  \bigskip
  \begin{center}
    {\LARGE\bf Title}
\end{center}
  \medskip
} \fi

\bigskip
\begin{abstract}
    Using representations of functional data can be more convenient and beneficial in subsequent statistical models than direct observations. These representations, in a lower-dimensional space, extract and compress information from individual curves. The existing representation learning approaches in functional data analysis usually use linear mapping in parallel to those from multivariate analysis, e.g., functional principal component analysis (FPCA). However, functions, as infinite-dimensional objects, sometimes have nonlinear structures that cannot be uncovered by linear mapping. Linear methods will be more overwhelmed given multivariate functional data. For that matter, this paper proposes a functional nonlinear learning (FunNoL) method to sufficiently represent multivariate functional data in a lower-dimensional feature space. Furthermore, we merge a classification model for enriching the ability of representations in predicting curve labels. Hence, representations from FunNoL can be used for both curve reconstruction and classification. Additionally, we have endowed the proposed model with the ability to address the missing observation problem as well as to further denoise observations. The resulting representations are robust to observations that are locally disturbed by uncontrollable random noises. We apply the proposed FunNoL method to several real data sets and show that FunNoL can achieve better classifications than FPCA, especially in the multivariate functional data setting. Simulation studies have shown that FunNoL provides satisfactory curve classification and reconstruction regardless of data sparsity.
\end{abstract}

\noindent%
\vfill

\newpage
\spacingset{1.5} 

\section{Introduction}\label{sec:intro}
Representation learning can be a powerful tool to extract features for functional data. Functional data are curves, images, or any objects defined over time, location, and other continuums \citep{fdabook, ferratybook, kokoszkabook}. Functional data are more prone to having nonlinear structures which would require substantial effort to process for further utilization. Directly using observed functional data for supervised models can be inefficient and even more so when observations are contaminated with uncontrollable random errors. Hence, we often use a low-dimensional latent feature space to represent the original functional data. The representations can be used for denoising functional data for recovering the individual functions. In addition, the representations can convey more information than interpolating the observed data. For example, the feature space may contain grouping or label information of observed trajectories. Building a classification model is thus more efficient by using such representations than original observations. The aforementioned matters motivate us to construct a representation learning model for functional data.    

Many representation learning methods have been developed for scalar variables, but there are only a few methods dedicated to functional data. The most used approach is the functional principal component analysis (FPCA), which aims to find major sources of variation in functional data. The major sources of variation are represented by the functional principal components (FPCs). Individual functional data can be expressed by a linear combination of FPCs. The coefficients to the FPCs are called FPC scores. The scores are the representations produced by FPCA, and they are linear projections in the low dimensional space defined by selected FPCs. We can then establish functional regression models based on FPC scores so that the model estimation is less burdened by the dimensionality of functional data. FPCA also provides an inherent way to reconstruct underlying functional data observed with random errors. \cite{fdabook} provided a comprehensive introduction on FPCA. The theories of FPCA were well established by \cite{fpca_asmp_3}, \cite{fpca_asmp_1}, and \cite{fpca_asmp_2}. Several extensions to standard FPCA have been proposed to adjust the representations for different scenarios. For instance, FPCs can be more interpretable when using a sparsity assumption as illustrated in \cite{localizedFPCA} and \cite{iFPCA}. Furthermore, \cite{rieFPCA} considered the case where functional data are defined on some Riemannian manifolds than the usual Euclidean space. As the representations of functional data, the FPC scores are used for subsequent statistical models. FPC scores can be used for establishing functional linear models~\citep{functional_linear} or generalized functional linear models~\citep{GFLM}. Moreover, the scores can be used for either clustering~\citep{Clustering_FPCA} or classification~\citep{Classification_fpca} of individual curves. Curve classifications by FPC scores were further investigated in \citet{perfectclassification} and \citet{centroidclassifier}. In real applications, functional data are not always ideal. \cite{sparseFPCA} and \citet{pcaforsparsefunctionaldata} addressed a common issue where functional data are not fully observed. In some cases, functional data contain excessive noises that would undermine the usual estimation process. \citet{kongdehan_outlier} and \citet{pcaforsparsefunctionaldata} demonstrated how to introduce robustness into the estimation for establishing a functional regression model or applying FPCA.

FPCA has been well developed for univariate functional data but cannot directly address multivariate data. Explaining the variation of multivariate trajectories becomes more difficult. Representing multivariate functional data by vectors are more beneficial for subsequent models in comparison to univariate case. \cite{fdabook} suggests one approach for dealing with multivariate functional data. That is, we can join multivariate trajectories in a univariate one, i.e., connecting functions by endpoints from all dimensions. The resulting functional data becomes univariate, and standard FPCA can then be applied to the obtained univariate functions. Different functional variables have different measurement units and magnitudes, hence \cite{chiouFPCA} employs normalization to balance the degrees of variations across dimensions. In the meantime, \cite{mfpca_differentdomain} investigated the case that trajectories from different dimensions are observed over non-uniform domains. 


Linear projections have at least the following limitations for representing functional data. The foremost limitation is that linear projection cannot recognize the nonlinear structures in functional data. Projections in the linear feature space may not offer more effective information compared with original observations. Second, FPCA lacks innate approaches for dealing with multivariate functional data. The conventional approach is to concatenate the multivariate variable into a univariate one and apply FPCA to the concatenation of functional data. As a result, the linear components have to consider the nonlinear structures compounded by all dimensions. The inadequacy of linear representations is more severe in the multivariate case. Thus, we need to pursue a mapping that addresses the nonlinearity in functional data and can be directly applied to multivariate data. The third limitation of FPCA is that it relies on the assumption of a common covariance function of all observed data. The associated functional principal space is a direct consequence of the assumption. However, if functional data contain labels for individual trajectories, then the assumption of a common covariance may be violated. The violation undermines the validity of the functional principal component space which does not convey label information. Alternatively, we can separately apply FPCA to observations with different labels. But the projections in a different functional principal component space are not comparable and cannot be used for developing a classification model. Therefore, we need to design a mapping that does not depend on the estimation of covariance functions. This mapping can be applied to observed data regardless of labels. The corresponding representations convey necessary label information and are comparable in a shared feature space. As a result, the representations can be passed onto the subsequent classification model. Last but not least, after obtaining the functional principal components, we usually keep the leading few FPCs and discard the rest because the leading FPCs contain the most information of functional data with respect to variations. Therefore, the final FPC space is a truncated space and may not be optimal for subsequent models. The low-rank FPCs may also be essential in the classification of functional data and sometimes more significant than the leading FPCs. This limitation motivates us to establish a mapping that does not discard any information in doing dimension reduction for functional data.  

In this work, we propose a functional nonlinear learning (FunNoL) method for functional data to overcome the aforementioned limitations of linear projections. The FunNoL method employs a nonlinear mapping $\phi: \mathcal{X} \rightarrow \mathcal{F}$ for functional data $\bm{X}(t)$ in the functional space $\mathcal{X}$ and establishes a feature space $\mathcal{F}$. Assuming that the dimension of the representation $\bm{z}$ in $\mathcal{F}$ is a much lower dimensional object than the original functional data, we can use $\phi$ to compress information of any given individual curve. The nonlinear mapping $\phi$ is not subject to any specific form. It can attend to nonlinear structures of functional data that would be overlooked by linear methods. We use the most popular tool, neural networks, to implement and estimate the nonlinear mapping. In real applications, functional data are always subject to excessive noises or missing observations. These can be thought of examples from the data corruption problem. For that matter, we introduce robustness to the representation model for dealing with missing observations and excessive local disturbances of observed functional data. The obtained representations can focus on information from the underlying function. 

Our proposed FunNoL method has at least the following advantages in comparison with FPCA. First, FunNoL addresses the nonlinear structures in functional data with a flexible nonlinear mapping estimated by neural networks. It alleviates the requirement of specifying any forms for the mapping functions. Second, multivariate functional data can be input to the networks without being changed to the univariate one. Third, FunNoL does not need to calculate covariance functions which require substantial computational efforts when the sample size or the number of observation points is large. Fourth, our proposed FunNoL method is a one-step model by which feature extraction and curve classification are done at the same time. The nonlinear mapping $\phi$ and its associated feature space are obtained under the criteria containing two objective functions for curve reconstruction and classification. Hence, the resulting feature space is optimal for both tasks. In comparison to the proposed FunNoL method, the conventional FPCA based approaches are two-step methods, i.e., they extract features (linear projections) in the first step and use them for the classification/clustering in the second step. Last but not the least, we do not need to truncate the feature space which is an essential step in FPCA, hence the representations can preserve as much information as possible. In conclusion, FunNoL can provide reliable and efficient low-dimensional representations of functional data.


We consider two modeling tasks, classification and reconstruction of functional data, upon obtaining the representations in real data applications. Based on the representations, we can consider a classifier $f$, a linear or nonlinear function, to produce labels for observed curves. Establishing a classification model on vector-valued representations is much easier than that on functions. It is also straightforward to extend the classification model for multivariate response variables.  The secondary task is to recover underlying individual curves given noisy and corrupted observations. FPCA offers an easy solution to this task given that individual curves are defined as a linear combination of FPCs. On the other hand, representations obtained through a nonlinear mapping require an inverse mapping $\psi$ to reconstruct the input curves. Acquiring $\psi$ is hence included in the estimation process. In the later section, we will illustrate how to use neural networks to simultaneously estimate $\phi$, $\psi$, and $f$. Obtaining $\phi$ under two objectives allows the feature space to contain and exchange information between two tasks. This can also help to solve the over-fitting problem where each modeling task acts as a regularizer on the other. As a result, the features have better generalization ability and are more task-oriented than using a single modeling task.

Real applications are carried out with focuses on the aforementioned two tasks. We will apply the FunNoL method on several data sets to fairly examine its performance. The data sets are retrieved from \cite{UCRArchive2018}. The primary assessment for the model is based on the label prediction of test data. One evaluation of representations is the expressiveness indicating whether the following model can easily utilize the information in the feature space. For that matter, we can consider using a simple logistic regression to obtain predicted labels. Based on the comparison of prediction performance, the proposed FunNoL method establishes a feature space that is more effective than the FPC space with the same number of dimensions. We have also evaluated the performance of the FunNoL method under various sparsity settings of training and test data. The FunNoL method provides accurate classification and reconstruction results that do not suffer from having missing observations. 

In this article, we demonstrate the flexibility and usefulness of the FunNoL method for functional data. It addresses the nonlinear structures of observed curves and handles multivariate variables more naturally in comparison with FPCA. The introduced robustness of representations can overcome various challenges caused by data corruption. It is evident that the FunNoL method can be involved in more applications from FDA. The remainder of this article is organized as follows. Section~\ref{sec:method} outlines the representation learning model. A bound of generalization errors is provided, and we will demonstrate that the smoothness of mapping functions is enforced with manually corrupted observations. Section~\ref{sec:estimation} illustrates the estimation process of nonlinear mappings. Applications on real data sets are examined in Section~\ref{sec:application}. Section~\ref{sec:simulation} contains simulation studies to examine the ability of FunNoL for dealing with data corruption problems. Section~\ref{sec:Conclusion} concludes the article. The application and simulation codes are published at \url{https://github.com/caojiguo/FunNoL}.

\section{Methodology}\label{sec:method}

Representation learning of functional data entails the construction of a mapping $\phi$, from the data space $\mathcal{X}$ to the associated feature space $\mathcal{F}$. In this work, we propose to establish a nonlinear mapping, more general and flexible, for producing representations of functional data. The proposed functional nonlinear learning (FunNoL) method includes three components: $\phi$ for learning features, $\psi$ for reconstructing functional data, and $f$ for classifying functional data. We will approximate these nonlinear functions with the following neural network implementation:
\begin{align}
    \mathbf{h}_{j} &= {g_{h}}(\mathbf{W}\mathbf{x}_{j} + \mathbf{U}\mathbf{h}_{j-1}) \text{\quad and \quad}
    \mathbf{z} =  {g_{z}}(\mathbf{V}\mathbf{h}_{J}), \nonumber\\
    \tilde{\mathbf{h}}_{j} &= {g_{\tilde{h}}}(\tilde{\mathbf{W}}\mathbf{z} + \tilde{\mathbf{U}}\tilde{\mathbf{h}}_{j-1}) \text{\quad and \quad} \tilde{\mathbf{x}}_{j} = {g_{x}}(\mathbf{G}\tilde{\mathbf{h}}_{j}), \nonumber\\
    \mathbf{y} &= {g_{y}}(\mathbf{M}\mathbf{z}). 
    \label{eq:SRLmodel}
\end{align}
All {$g$}'s are activation functions to be chosen in applications. 
$\bm{W}$, $\bm{U}$, $\bm{V}$, $\tilde{\bm{W}}$, $\tilde{\bm{U}}$, $\bm{M}$, and $\bm{G}$ are parameter matrices of the recurrent neural networks. Let $\mathbf{x}_{j}$ be a $D$-dimensional vector corresponding to the evaluation of $\mathbf{X}(t_{j}) = (X_{1}(t_{j}),..., X_{D}(t_{j})$ at $t_{j}$. Let the hidden state $\mathbf{h}_{j}$ and the feature space be $L$-dimensional, and the curve label $\mathbf{y}$ is a $Q$-dimensional vector of binary values, such that the element in $\mathbf{y}$ is 1 for the observed label and zero otherwise. Then, $\mathbf{W}$ is a $L$ by $D$ matrix, and $\mathbf{U}$ is a $L$ by $L$ matrix. All of  $\mathbf{V}$, $\tilde{\mathbf{W}}$, and $\tilde{\mathbf{U}}$ are $L$ by $L$ matrix, whereas $\mathbf{G}$ and $\mathbf{M}$ are $L$ by $D$ and $L$ by $Q$ matrices.

\begin{figure}[H]
    \centering
    \includegraphics[width=\textwidth]{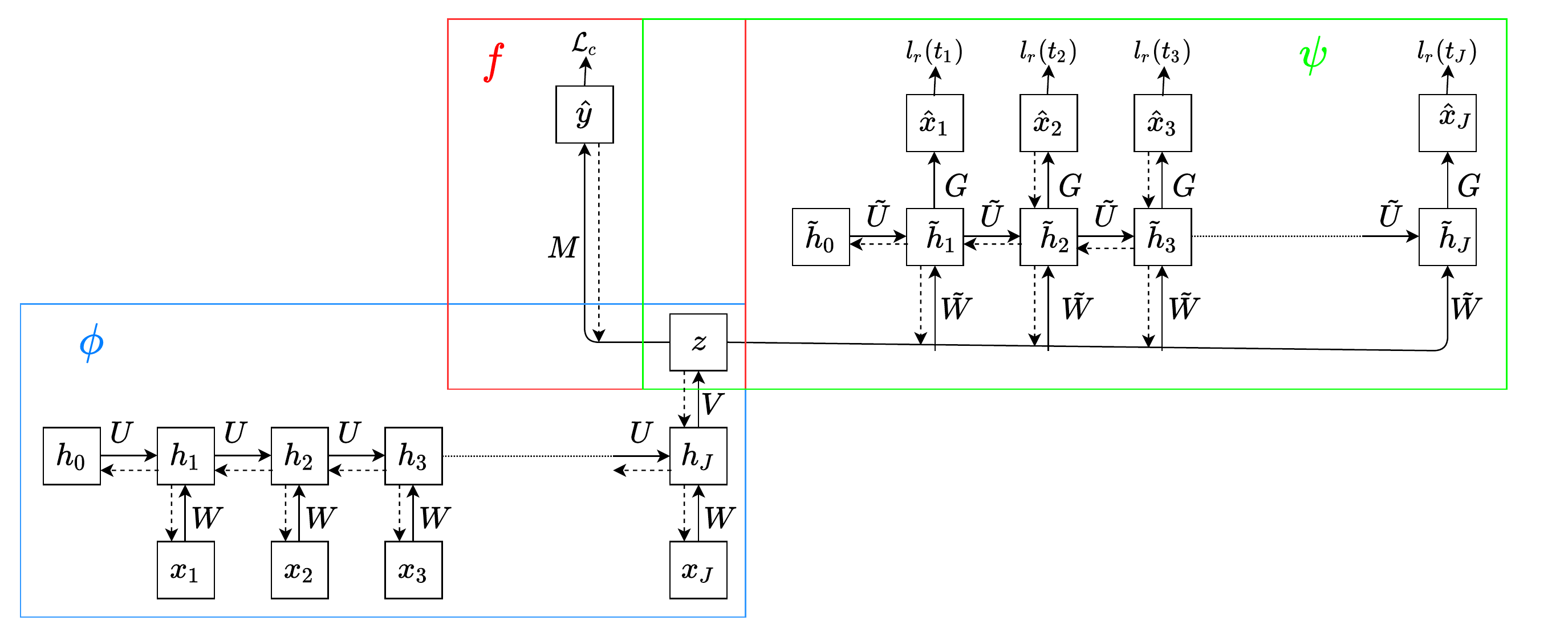}
    \caption{\label{fig:networkarch} The network design of the proposed functional nonlinear learning (FunNoL) model (\ref{eq:SRLmodel}). It includes three components $\phi$, $\psi$, and $f$ for learning features,  reconstructing functional data, and classifying functional data, respectively. The bold arrow lines correspond to the direction of propagation, whereas the dashed arrow lines indicate the flow of gradients. Boxes enclose the networks estimating each component in the model.}
\end{figure}

Figure~\ref{fig:networkarch} depicts the neural networks which estimate the representation model including the aforementioned three components. 

Explicitly defining a nonlinear mapping can be aimless in searching for the most optimal one. For that matter, we can use neural networks to approximate nonlinear functions. This allows us to obtain mappings for functional data without specifying any particular form for them. Another advantage of neural networks is that we do not need to process data before doing the estimation. Multivariate functional data can be directly fed into the network without being flattened into univariate trajectories. In the latter part of the section, we can also demonstrate that neural networks can handle missing observation, or data corruption, in functional data.

First, we will focus on the estimation of $\phi$ which takes the input of trajectories and produces a representation vector. Functional data is distinct from multivariate data where the former is defined over a continuum. The dependence between values at different observations points prohibits us to use standard feed-forward neural networks (FNN). If we input functional data into FNN, then the dependency in trajectories gets lost after the first layer where we take the inner product between them and coefficients. To be more adequate for functional data, we can use recurrent neural networks (RNN) which can maintain the continuity in observed trajectories. 

Let $\mathbf{X}(t)$ be a $D$-dimensional functional variable. We assume that the variable is observed over a grid of equally spaced points $t_{1},..., t_{J}$ for all individuals. Then, any observed trajectory is denoted by the sequence $\mathbf{x}_{j} = \{\mathbf{x}(t_{j})\} = (X_{1}(t_{j}),..., X_{D}(t_{j}))$ for $j=1,\ldots,J$. For simplicity of notation, we assume that all $D$ functional variables are observed at the same time, but our method can apply for the case when the functional variables are observed at different times.  One significant component of the RNN model is the hidden unit $\mathbf{h}(t)$ introduced to yield nonlinearity and abstraction in the network output. The hidden unit will be evaluated in parallel to $\mathbf{X}(t)$ at every observation point $t_{j}$. The role of $\mathbf{h}(t)$ is to create abstract dimensions to perform nonlinear transformation on observations $\mathbf{x}(t)$ along $t$. It will also use a nonlinear function to propagate hidden information from $\mathbf{h}_{j-1} = \mathbf{h}(t_{j-1})$ to $\mathbf{h}_{j} = \mathbf{h}(t_{j})$. We will use a general specification of a RNN to demonstrate how $\phi(\mathbf{X}(t))$ can be approximated. 

A simple RNN model is defined as $\mathbf{h}_{j} = {g_{h}}(\mathbf{W}\mathbf{x}_{j} + \mathbf{U}\mathbf{h}_{j-1})$ where $g_{h}$ is the activation. The hidden unit $\mathbf{h}$ is called the memory unit in the network. As illustrated in the above formulation, the hidden unit is responsible for handling the dependence between observations with a nonlinear autoregressive form. This also gives $\mathbf{h}_{j}$ the ability to memorize information up to $t_{j}$. In addition to tracking past information, the hidden unit uses abstract dimensions to process observations at each observation point. We can also observe that $\mathbf{h}$ is able to take multivariate observations. The dimension of $\mathbf{h}$, as the dimension of the feature space, will be treated as a tuning parameter. At the end, any observed trajectory can be summarized by $\mathbf{h}_{J}$. Applying a nonlinear mapping $\phi$ on any observed $\mathbf{x}(t)$ can be achieved by using $\mathbf{h}_{J}$ at an intermediate step. We generate the representation $\mathbf{z}$ as a function of $\mathbf{h}_{J}$ which can be approximated by a feed-forward neural network, i.e., $\mathbf{z} = g_{z}(\mathbf{V}\mathbf{h}_{J})$. Thus, the approximation of $\phi$ is based on the combination of two network models, from $\mathbf{x}$ to $\mathbf{h}_{J}$ and  $\mathbf{h}_{J}$ to $\mathbf{z}$.

The first objective upon obtaining the representations is to reconstruct and denoise functional data. The proposed nonlinear representation model requires us to establish an inverse mapping $\psi$ from the feature space to the original functional data space for completing the objective. Similar to the case of estimating $\phi$, we propose to use RNNs for the approximation of $\psi$ without defining an explicit form. The RNN model for approximating $\psi$ is specified as $\tilde{\mathbf{h}}_{j} = g_{\tilde{h}}(\tilde{\mathbf{W}}\mathbf{z} + \tilde{\mathbf{U}}\tilde{\mathbf{h}}_{j-1})$ and $
    \tilde{\mathbf{x}}_{j} = g_{x}(\mathbf{G}\tilde{\mathbf{h}}_{j})$ 
where $\tilde{\mathbf{x}}_{j}$ is the reconstruction at any given point $t_{j}$.  

The second objective after obtaining the representations $\mathbf{z}$ is to predict the label $\mathbf{y}$, in one-hot format, for functional data. We establish a classification model based on the representation, i.e., $\mathbf{y} = f(\mathbf{z})$. To align with the approximation of $\phi$ and $\psi$, we can use a feedforward neural network to estimate this nonlinear function, i.e., $\mathbf{y} = {g_{y}}(\mathbf{M}\mathbf{z})$ where $g_{y}$ is an activation function and $\mathbf{M}$ is the coefficient matrix. We will illustrate how to stack neural networks for approximating complex $\phi$, $\psi$, and $f$ in the supplementary document. In the remainder of this section, we provide the generalization ability of this model on predicting labels.


One concern of using neural networks for approximating nonlinear functions is the generalization ability. Given that the proposed model~(\ref{eq:SRLmodel}) is trained on one data set, whether it can be generalized for any unseen data is of great interest. That is, we expect the representation model can provide reliable predictions in all three model objectives. This concern is partially caused by the over-parameterization in the neural networks. For example when we are inputting functional data into the model, the first RNN component introduces a hidden unit $\mathbf{h}_{j}$ in parallel to any observed trajectories. The dimension of $\mathbf{h}$ is usually larger than the dimension $D$ in functional data. The correlated dimensions in $\mathbf{h}$ are used to extract as much information as possible. As we increase the dimension in the hidden unit, the dimensions of both coefficient matrices $\mathbf{W}$ and $\mathbf{U}$ are also increasing. The representation model and the classifier tends to over-fit the training data given the increasing model complexity as well as more stacked layers of networks.

To address such concern, we need to quantify the generalization ability of the proposed representation model. We can use the generalization error which corresponds to the expected model errors over a data distribution. By providing a bound on the generalization error, we may have a better understanding of prediction performances with repsect to the proposed method. In this work, we will prioritize the classification task over the reconstruction task and develop a generalization bound for it. Let the classification loss function $l_{c}$ to be the margin loss, and the expected margin losses given a data distribution is given in the following definition.
\begin{definition}
\texttt{Expected margin loss}. Let $\mathcal{D}$ be the distribution of any data pair $(\bm{x},\bm{y})$ and $\gamma$ corresponds to the margin in classification. Then, the expected margin loss with respect to $\mathcal{D}$ is 
\begin{equation}
    l_{\gamma}(f_{\bm{w}}) = \mathbb{P}_{(x,y) \sim \mathcal{D}} [ f_{\bm{w}}(\bm{x})[y] \leq \gamma + \text{max}_{j \neq y}\{f_{\bm{w}}(\bm{x})[j]\}],
    \label{eq:marginloss}
\end{equation}
\end{definition}
\noindent where $f_{\bm{w}}(\bm{x})[y]$ is the prediction of class $y$ in $\bm{y}$ and $\mathbf{w} = (\bm{W}, \bm{U},\bm{V}, \tilde{\bm{W}}, \tilde{\bm{U}}, \bm{M}, \bm{G})$ is the parameter vector. 
Let $f_{\mathbf{w}}(\mathbf{x}) : \mathcal{X} \rightarrow \mathbb{R}^{Q}$ be the classifier based on some covariates $\mathbf{x} \in \mathcal{X}$ and $Q$ is the number of classes in $\mathbf{y}$. The curve label $\mathbf{y}$ is a $Q$-dimensional vector of binary values, such that the element in $\mathbf{y}$ is 1 for the observed label and zero other wise. The output of $f_{\mathbf{w}}(\mathbf{x})$ is also a $Q$-dimensional vector. Each element in $f_{\mathbf{w}}(\mathbf{x})$ is the probability of $\mathbf{x}$ belonging to the corresponding class in $\mathbf{y}$.
Given any training data set $\mathcal{S}$, the empirical margin loss is calculated as 
\begin{equation*}
    \hat{l}_{\gamma}(f_{\bm{w}}) = \frac{1}{|S|}\sum_{(x,y) \sim S} [f_{\bm{w}}(\bm{x})[y] \leq \gamma + \text{max}_{j \neq y}\{f_{\bm{w}}(\bm{x})[j]\}].
\end{equation*}
We can get the generalization bound on the classifier in model~(\ref{eq:SRLmodel}) as follows.
\begin{theorem}
(Generalization bound). For any $\delta, \gamma > 0 $, with probability $\geq 1 - \delta$ over a set of $N$ data pairs, the following bound holds
\begin{equation*}
    l_{0}(f_{\bm{w}}) \leq \hat{l}_{\gamma}(f_{\bm{w}}) + \mathcal{O}(\sqrt{\frac{E(\bm{w}) + \text{ln}({N}/{\delta})}{N-1}}),
\end{equation*}
where $E(\bm{w}) = \frac{(H_{y,M} + H_{y,V} + H_{y,W} +H_{y,U})k\text{ln}k}{\gamma^{2}}(\frac{||\bm{M}||^{2}}{||\bm{M}||^{2}_{F}} + \frac{||\bm{V}||^{2}}{||\bm{V}||^{2}_{F}} + \frac{||\bm{W}||^{2}}{||\bm{W}||^{2}_{F}} + \frac{||\bm{U}||^{2}}{||\bm{U}||^{2}_{\mathcal{F}}})$,
\label{thm:gb}
\end{theorem}
\noindent For Theorem~\ref{thm:gb}, we assume that the dimension of $\bm{h}$ and number of variables in $\bm{X}(t)$ are the same as $k$.The $||\cdot||$ is the Euclidean norm, and $||\cdot||_{F}$ is the Forbenius norm such that $||\mathbf{A}||_{F} = \sqrt{\sum_{i=1}^{m}\sum_{j=1}^{m}A_{ij}^{2}}$ for a $m$ by $m$ matrix $\mathbf{A}$. The proof of the theorem is presented in the supplementary document. 

The next section will illustrate how to estimate parameters of the networks. We also address an optimization obstacle, exploding/vanishing gradient, and provide solutions to the problem. Given that the observed trajectories may contain excessive noises, we introduce robustness into the estimation routine so that the representations are insensitive to local disturbances on the observations.

\section{Model Estimation}\label{sec:estimation} 
The neural network parameters in $\bm{w} = (\bm{W}, \bm{U}, \bm{V}, \bm{M}, \tilde{\bm{W}}, \tilde{\bm{U}}, \bm{G})$ are estimated by minimizing the following objective function:
\begin{align}
    \hat{\bm{w}} &= \argmin_{\mathbf{w}}\mathcal{L} =  \argmin_{\mathbf{w}} \mathcal{L}_{c} + \mathcal{L}_{r},\nonumber \\ 
   &=\argmin_{\mathbf{w}}\frac{1}{N}\sum_{i=1}^{N} l_{c}(y_{i}, f(\bm{z}_{i})) +   \frac{1}{N}\sum_{i=1}^{N} l_{r}(x^{i}_{1:J}, \psi(\phi(x^{i}_{1:J}))),
    \label{eq:trainingobj}
\end{align}
where $l_{c}$ and $l_{r}$ are the loss functions for assessing the classification and reconstruction results. The loss function $l_{r}$ needs to be decomposed with respect to $t_{j}$, i.e., $l_{r}(x^{i}_{1:J}, \psi(\phi(x^{i}_{1:J})))  = \frac{1}{J}\sum_{j=1}^{J}l^{(i)}_{r}(t_{j})$. For the example of using a squared error loss function, $ l^{(i)}_{r}(t_{j})=  ||\mathbf{x}^{i}_{j} - \tilde{\mathbf{x}}^{i}_{j}||^{2}$. Then, the optimization in~(\ref{eq:trainingobj}) is done by updating parameters with $\bm{w} \leftarrow \bm{w} - \alpha \cdot {d\mathcal{L}}/{d\bm{w}}$ for a learning rate $\alpha$. For illustrating the estimation procedure, we will use the original observations. Adding a corruption step does not alter the derivation of parameter gradients. In case of missing observations, we can set the loss function to be zero at their observed locations.

The gradient update is also termed the backpropogation under the framework of neural networks. The gradients of parameters are calculated in the opposite direction of the network flow. The dashed lines in Figure~\ref{fig:networkarch} are showing the order of calculations. It is easy to backpropogate training errors in the standard feed-forward neural network which is estimating $f$. On the other hand, updating parameters in recurrent neural networks are more complicated. The training error is not only transferred through layers but time $t_{j}$'s as well. Backpropogation through time (BPTT) is involved with the estimation of $\phi$ and $\psi$. Figure~\ref{fig:networkarch} illustrates both the network flow and direction of backpropogation. The detailed derivations of gradients are included in the supplementary document.

\subsection{Vanishing/exploding gradient}
The recurrent neural network and its variants are more prone to the vanishing or exploding gradient problem. That is, the gradients become too small or huge when applying updates on the current parameter estimates. After updating, the estimates are far away from the previous ones, and the optimization procedure becomes unstable. The issue is primarily caused by applying the chain rule over time, that is, 
\begin{equation*}
    \frac{d\bm{h}_{j}}{d\bm{h}_{s}} = \prod_{k=s+1}^{j}\frac{d\bm{h}_{k}}{d\bm{h}_{k-1}} = \prod_{k=s+1}^{j} \bm{U}\text{diag}(\sigma_{h}^{\prime}(\bm{W}\bm{x}_{k} + \bm{U}\bm{h}_{k-1})).
\end{equation*}
The norm of $d\mathbf{h}_{k}/d\mathbf{h}_{k-1}$ is bounded by the norms of both terms in the product. As $j-s$ increases, $||d\mathbf{h}_{k}/d\mathbf{h}_{k-1}|| < 1$ leads to the vanishing gradients whereas $\||d\mathbf{h}_{k}/d\mathbf{h}_{k-1}|| >1 $ leads to the exploding case. This recursive term is unavoidable in recurrent networks and a major performance setback to estimations. The root of this problem is embedded in the long-term dependency between hidden units $\bm{h}_{j}$ and $\bm{h}_{s}$. From the specification in~(\ref{eq:SRLmodel}), $\bm{h}_{j}$ congregates all information from $\bm{h}_{s}$ for any $s < j$ regardless of the history significance. Given that we have a constant coefficient matrix $\mathbf{U}$ for weighting the previous hidden state, the product in $d\bm{h}_{j}/d\bm{h}_{s}$ is the result of multiplying the same coefficient over and over. This will lead to the gradient problem. One naive solution is to truncate the history dependence within a fixed threshold distance. If $j-s$ exceeds the threshold, we only calculate the recursive gradient for a fixed number of times. This approach fails to attend to important relations persisting through the distance in $t$. The rest of this section introduces remedies for the problem which do not truncate the history dependence.

The first approach to address the issue is the gradient clipping method. The clipping strategy is primarily focusing on the exploding gradient scenario (\citet{gradientclipping}) and can attend to the vanishing gradient problem with some modifications. The logic of this approach utilizes unstable gradients. Even though the gradient is either exploding or vanishing, it still points to the right direction of optimization. As a result, we need to control them from making the optimization procedure unstable. The clipping method is to impose a threshold on the norm of the gradient. If the norm of the gradient is larger than the threshold, then we scale down the gradient to a unit-norm vector.


Alternatively, it is more intuitive to enhance the hidden states with abilities to handle history dependence. In other words, the history dependence should be varying and determined by the model. The main cause for gradient problems is from the constant coefficients $\mathbf{U}$ and $\tilde{\mathbf{U}}$ in~(\ref{eq:SRLmodel}). These describe a constant dependence between adjacent hidden states throughout the entire observation period. The motivation of this approach lies in replacing a constant coefficient matrix with a varying quantity. The solution is to introduce an additional hidden unit $\mathbf{m}$ as well as gating functions to process information from past hidden states. One popular model following this approach is the long short-term memory model (LSTM) as the following: 

\begin{align}
    \mathbf{i}_{j} &= g(\mathbf{W}_{i}\mathbf{x}_{j} + \mathbf{U}_{i}\mathbf{h}_{j-1}), 
    \mathbf{o}_{j} = g(\mathbf{W}_{o}\mathbf{x}_{j} + \mathbf{U}_{o}\mathbf{h}_{j-1}), 
    \mathbf{f}_{j} = g(\mathbf{W}_{f}\mathbf{x}_{j} + \mathbf{U}_{f}\mathbf{h}_{j-1}), \nonumber\\
    \mathbf{m}_{j} &= \mathbf{f}_{j} \circ \mathbf{m}_{j-1} + \mathbf{i}_{j} \circ \tilde{\mathbf{m}}_{j} \text{ where }\tilde{\mathbf{m}}_{j} = \text{tanh}(\mathbf{W}_{m}\mathbf{x}_{j} + \mathbf{U}_{m}\mathbf{h}_{j-1}), \nonumber\\
    \mathbf{h}_{j} &= \mathbf{o}_{j}\circ\text{tanh}(\mathbf{m}_{j}),
    \label{eq:LSTM}
\end{align}
where $g$ is the sigmoid function. LSTM defines three gates: the input gate $\mathbf{i}(t)$, the forget gate $\mathbf{f}(t)$ and the output gate $\mathbf{o}(t)$. These gating functions are bounded within $[0,1]$ and act as a weighting function dependent on the observation point. They are responsible for handling external inputs and past information, updating the memory in $\bm{m}(t)$, and producing the current hidden state $\bm{h}(t)$. LSTM facilitates the propagation from $\bm{h}_{j-1}$ to $\bm{h}_{j}$ by $\bm{m}_{j}$ instead of direct connection between adjacent states. This replaces the recursive component from $d\bm{h}_{j}/ d\bm{h}_{j-1}$ with $d\bm{m}_{j}/d\bm{m}_{j-1}$ in the calculation of gradients through time. The gating functions are present in the calculation of $\bm{m}_{j}$. Since the gating functions are varying for reflecting the temporal relationship, they can prevent the recursive gradient to use a constant value. As a result, LSTM and its variants can avoid the vanishing/exploding gradient problem. To illustrate the usefulness of LSTM model, we can look at the derivative $dm_{j}/dm_{j-1}$ which is derived in the supplementary document. In real applications, we will use the LSTM model for obtaining representations of functions.

\subsection{Robustness}\label{ssec:robustness}
Robustness is crucial for establishing a representation model. It facilitates the proposed model in real applications where data is not always ideal. We would encounter several data problems preventing a fitted model to be useful in doing predictions. First, observed trajectories are often contaminated by random errors which do not have the same variance nor follow a normal distribution. At some locations, trajectories can be departed from the underlying function by outliers. We can also have random missing observations which forbid us from directly fitting any model. These data problems will become more severe when they are present across all dimensions. One approach for addressing these problems is to smooth functional data before constructing a prediction model. This would separate a single representation model into two and introduce additional estimation burdens. It is also difficult to verify the validity of smoothed functional data. As a result, we indulge the proposed model with robustness so that it can be insensitive to data corruption. This can help the obtained representations be less prone to uncontrollable errors and more reliable for further modeling tasks.

The alternative view of this issue is how to enforce the smoothness of the representation mapping $\phi$. One important difference between curves and infinite-dimensional vectors is the smoothness assumption. That is, any random trajectory should behave in the way such that $\bm{x}(s) \approx \bm{x}(t)$ given $s \approx t$. If we can ensure the smoothness of reconstructed curves $\psi(\phi(\mathbf{x}(t)))$, then we are able to ensure that both $\phi$ and $\psi$ are smooth. Traditional methods from FDA will impose some analytic roughness penalties to enforce the smoothness of curves. In our method, we do not express the mapping function as a linear combination of basis functions, the classic FDA treatment. Therefore, we cannot directly obtain the roughness penalty for the mapping functions. The roughness penalty is usually computed by integration. Numeric approximation to such integration depends on both the number of observations points and the completeness of any trajectory. Hence, adding an analytic roughness penalty puts too much burden on the estimation process. 

Following the penalizing strategy, we propose to modify the estimation procedure for approximately enforcing the smoothness of mapping functions. This will also ensure the robustness of the proposed representation model. The modification is to introduce a corruption layer on functional data. We will show that this is equivalent to adding an approximate roughness penalty on the reconstructed trajectories. By penalizing the roughness of reconstructions, we can consequently control the smoothness of $\phi$ and $\psi$. As a result, the representations will become insensitive to random noises and missing observations. Training neural networks with added noises demonstrates the ability to enforce smoothness of mapping functions for scalar variables~\citep{trainingwithnoise}. In this work, we will demonstrate that this modified estimation will provide robustness for the model~(\ref{eq:SRLmodel}) and subsequently more reliable representations for further modeling tasks.

We will introduce a corruption layer to produce corrupted data before fitting the model. The representation model is then estimated with corrupted inputs while providing reconstruction and classification. The corruption process is a combination of discarding observations and adding random noises for imitating problems in real data. Let $p(x^{\star}_{1:J}|x_{1:J})$ represents a stochastic mapping for producing the corrupted input $x^{\star}_{1:J}$. The corruption step $p(x^{\star}_{1:J}|x_{1:J})$ can be adding uncorrelated Gaussian errors to $x_{1:J}$, or it can randomly destroy individual elements in $x_{1:J}$ as if they are missing from the observed sequence. For illustration, we will first randomly discard observations and set them to be missing. Then, we will use $x^{\star}_{j} = x_{j} + \epsilon_{j}$ for obtaining corrupted input given that $x_{j}$ is not missing. The Gaussian errors $\epsilon_{j}, j=1,\ldots,J$ are i.i.d and have a variance of $\sigma^{2}$. We will impose the same corruption step to each dimension if $\bm{x}(t)$ is not univariate. As a result, the representation model is instead obtaining features through $\phi(\bm{x}^{\star}_{1:J})$ based on the corrupted input and alternating the reconstruction loss as $\frac{1}{N}\sum_{i=1}^{N}l_{r}(\bm{x}^{i}, \psi(\phi(\bm{x}^{i,\star})))$. Using $\psi(\phi(\bm{x}^{i,\star}))$ in $l_{r}$ will enrich the representations with robustness for better denoising capabilities as well as reconstructions based on partially observed inputs. This training strategy is equivalent to adding an approximate roughness penalty on the mapping functions. In the online supplementary document, we demonstrate the correspondence between the modified training process and the roughness penalization of mapping functions. 

\section{Applications}\label{sec:application}
In this section, we will present the real applications of the FunNoL method to illustrate its performance on curve classification. We select multiple data sets from~\cite{UCRArchive2018} and carry out FPCA on each data set for making comparisons. The comparison results on the curve reconstruction are presented in Section S7.3 of the supplementary document. 

\subsection{Applications on univariate functional data}
First, we will apply the FunNoL method to univariate functional data. Descriptions and illustrations of the data sets are included in Section S7.1 of the supplementary document. The applications are carried out in the following scheme. For each data set, we split the data into training and test set. The FunNoL method and FPCA are first estimated based on the training set. We will use corrupted and original observations for fitting the FunNoL model (denoted by FunNoL\_c and FunNoL\_nc respectively). Then, we use the estimated models, both FunNoL and FPCA, to predict representations of observations in the test set. 

The main assessment for representations is focused on their efficacy in classification. We will use the prediction accuracy on labels from test data for demonstrating the classification performance. The scheme of obtaining the prediction accuracy is as follows. We first use the representations, obtained from the estimated models, to establish a logistic regression model for fitting the labels in the training set. After predicting representations for trajectories in the test set, the logistic model produces labels based on the predicted representations. For each data set, we will split it into training and test set and repeat data splitting 50 times. For each splitting, we will repeat the estimation and prediction procedure to get the classification accuracy on the test set. The prediction performances from the 50 random splits are recorded for making comparisons. 

We apply the same scheme for both FunNoL and FPCA and compare the portion of corrected labels between these two approaches over all splits. The classification results on the test sets are summarized in Table~\ref{tab:1D_classification_full}. For making comparisons, we fix the number of features from FunNoL to be the same as the number of FPCs. Table~\ref{tab:1D_classification_full} shows that adding a corruption step improves the prediction accuracy across data sets with different training sample sizes. The improvement is most significant when the sample size is large from the data set \texttt{Starlight}. Thus, we can observe that manually corrupting data can prevent the FunNoL model from overfitting the training data and improve the generalization ability of representations. In comparison with FPCA, FunNoL is better in classification when the training sample size is large. The nonlinear mappings can be well approximated by neural networks with a large number of samples. When the training sample size is small, it is not clear which method produces better representations for classification.

\begin{table}[H]
    \caption{\label{tab:1D_classification_full}Prediction accuracy from FunNoL (with or without the corruption step, denoted by FunNoL\_c and FunNoL\_nc respectively) and FPCA. The first row indicates the training sample size for each data set. Each cell in the last three rows provides the classification accuracy with standard errors in parentheses on curve labels in the test sets.}
\centering
    \begin{tabular}{c|c|c|c|c}
        \hline 
          & \texttt{Starlight} & \texttt{Wafer}   &  \texttt{SwedishLeaf} & \texttt{Earthquakes} \\
          \hline 
         \# Training samples & 8236 & 6164 &  625  & 322 \\
        \hline 
        FunNoL\_c & .975(.003) & .995(.002)  &  .398(.052) & .839(.035)\\
        \hline 
        FunNoL\_nc & .951(.003) & .995(.002)  &  .381(.042) & .814(.027) \\
        \hline 
        FPCA & .847(.006) & .927(.005) &  .471(.057)  & .810(.032) \\
        \hline 
\end{tabular}

\end{table}

\begin{figure}[H]
    \centering
    \includegraphics[width=.49\textwidth]{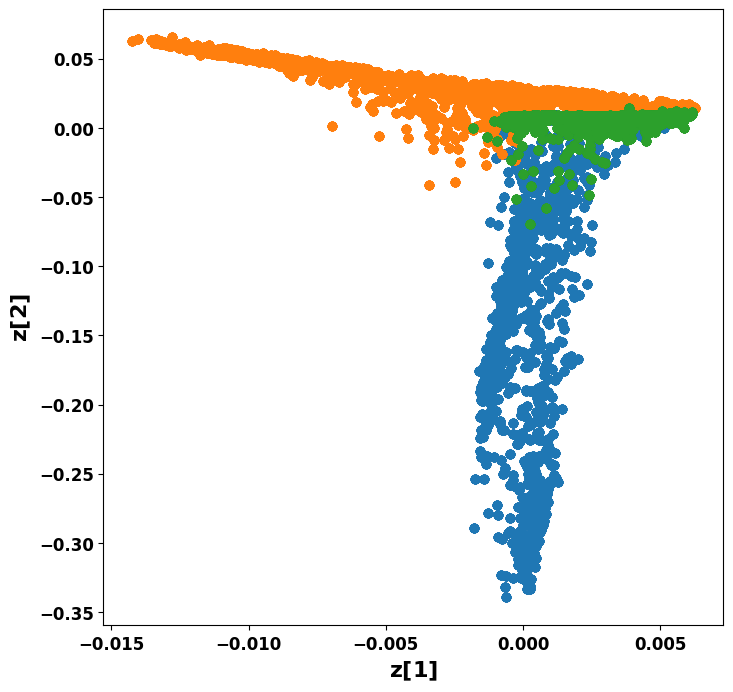}
    \includegraphics[width=.49\textwidth]{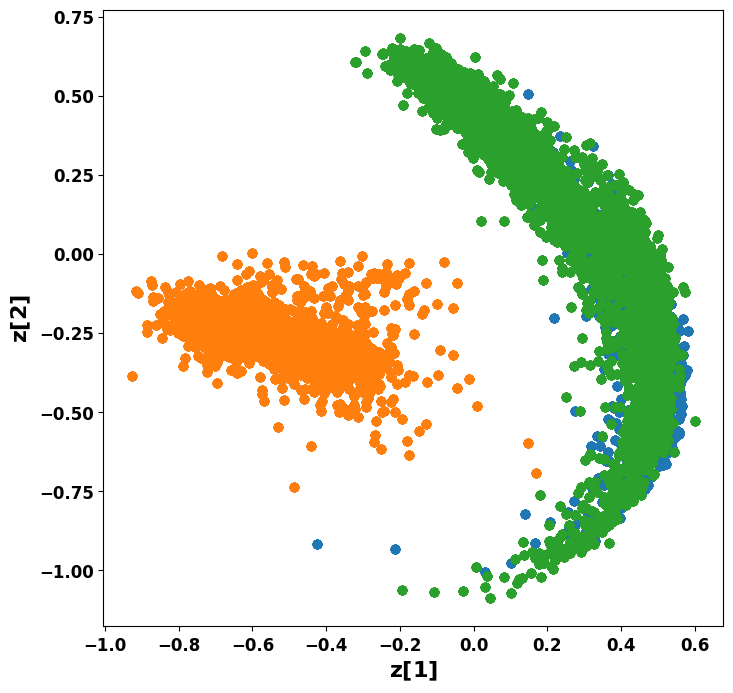}
    \caption{\label{fig:starlight_latent}The first two dimensions from the feature space of FunNoL (left) and the two leading functional principal dimensions of FPCA (right) for the \texttt{Starlight} data set. Colors correspond to the classes of individual trajectories.}
\end{figure}

For further investigation of FunNoL, we can examine the latent feature space and compare it with the space spanned by the leading FPCs. This can also helps us to understand how representations provide information for classification. Figure~\ref{fig:starlight_latent} presents the first two latent features from FunNoL\_c and two leading FPC scores for the \texttt{Starlight} data set. In this data set, we have two classes (plotted in green and blue curves at the top left panel of Figure S1 in the supplementary document) that have similar trajectories. Their projections in the FPC space are not easily separable with a linear classifier. On the other hand, applying a nonlinear mapping produces representations which are showing better separation between these two classes. In conclusion, FunNoL is able to produce representations which are better suited for classification of univariate functional data in comparison with FPC scores.

\subsection{Applications on multivariate functional data}
\begin{figure}[H]
\centering
\includegraphics[width=.49\textwidth]{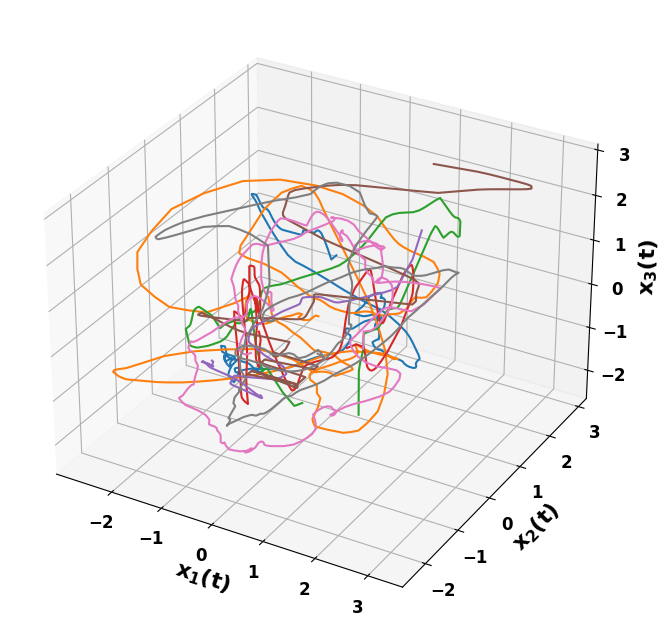}
\includegraphics[width=.49\textwidth]{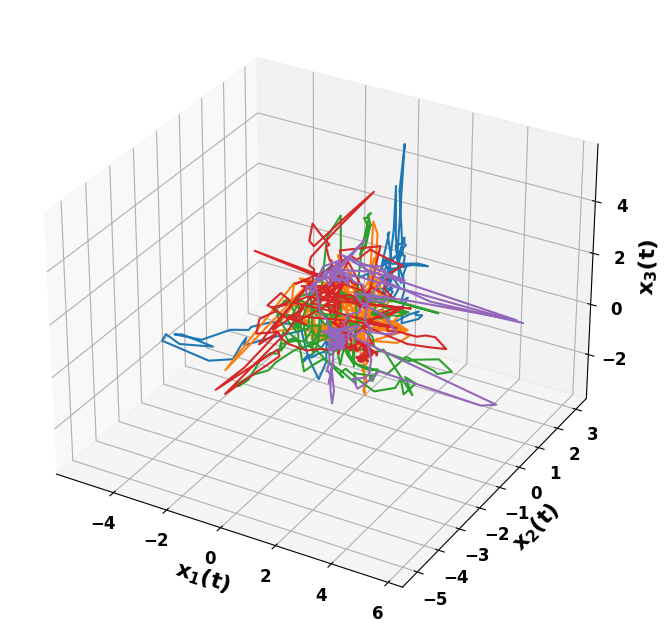}
\caption{\label{fig:gesture_showcase} A few selected trajectories from the \texttt{Gesture} and \texttt{Cricket} data set. In both graphs, each color indicates a different class of trajectory recorded in a three dimensional space over time.} 
\end{figure}

One motivation of this work is to focus on multivariate functional data that are more common in applications. To show the advantages of FunNoL in learning multivariate functional data, we have applied it to two data sets \texttt{Gesture} and \texttt{Cricket} from~\cite{UCRArchive2018}. Figure~\ref{fig:gesture_showcase} depicts sample trajectories from both the \texttt{Gesture} and \texttt{Cricket} data set.

One advantage of FunNoL is that it can be directly applied to multivariate functional data. This makes it more suitable for addressing the nonlinear structures of multivariate functions without concatenating them into univariate ones. We will repeat the splitting routine as done to the previous univariate functional data. For each split, the prediction accuracy on curve labels is recorded and summarized in Table~\ref{tab:3D_classification_full}.
\begin{table}[H]
    \caption{\label{tab:3D_classification_full}Prediction accuracy from FunNoL (with or without the corruption step, denoted by FunNoL\_c and FunNoL\_nc respectively) and FPCA on multivariate functional data. mFPCA is to obtain FPC scores by the multivariate FPCA method ~\citep{mfpca_differentdomain}. The first row indicates the training sample size for each data set. Each cell in the last three rows provides the classification accuracy with standard errors in parentheses on curve labels in the test sets.}
\centering
\begin{tabular}{c|c|c}
    \hline 
      & \texttt{Gesture} & \texttt{Cricket}  \\
      \hline 
     \# Training samples & 3582 & 390\\
    \hline 
    FunNoL\_c & .963(.0044) & .316(.031)  \\
    \hline 
    FunNoL\_nc & .920(.0055) & .368(.033)  \\
    \hline 
    FPCA & .880(.0080) & .0791(.023) \\
    \hline 
    mFPCA & .887(.0076) & .0842(.021) \\
    \hline
\end{tabular}

\end{table}
We observe a more significant improvement in the prediction accuracy than the case of univariate functional functional data. The nonlinear representations provide a better ground to build classification models for multivariate functional data. FunNoL outperforms FPCA even with small sample size as in the \texttt{Cricket} data set. Adding a corruption step does not necessarily lead to an increase in the prediction accuracy. From Tabel~\ref{tab:3D_classification_full}, the influence of a corruption step depends on the training size. Data sets with more samples (as in the \texttt{Gesture} data set) can benefit from corrupting data for training representation models. We will examine the classification of multivariate functional data in the next section by simulation studies.

\section{Simulations}\label{sec:simulation}
This section includes simulation studies for assessing the FunNoL method. We examine whether data corruption impairs the performance of FunNoL on the aforementioned modeling tasks. These studies can also be used for verifying the robustness of the proposed method. In real applications, functional data are not always ideal where trajectories may not be fully observed. That is, each observed trajectory may have missing observations. The observations may also contain excessive noises disturbing the resulting representations.

As stated in the previous section, FunNoL uses a corruption step in estimation to reduce influences from corruption in functional data. In the simulation studies, we are aiming to show whether representations from FunNoL are reliable in modeling tasks given different levels of corruption in data. For that matter, we investigate whether the portion of missing observations in trajectories undermines the reliability of the learning model. 


Data sets presented in the previous section are used in the simulations. Let the model estimated by the complete data be the benchmark. We will gradually decrease the percentage of available data from 90\% to 10\% for implementing FunNoL and FPCA. For each setting, we will take a random sample from each trajectory with 90\%, \dots, 10\% of the complete data, and set the remaining to be missing. Given the downsampled data, we will repeat data splitting and model estimation 50 times. The prediction accuracy on labels in test sets is obtained as the previous applications and recorded.

\subsection{Simulation on univariate functional data}
\begin{figure}[H]
    \includegraphics[width=\textwidth]{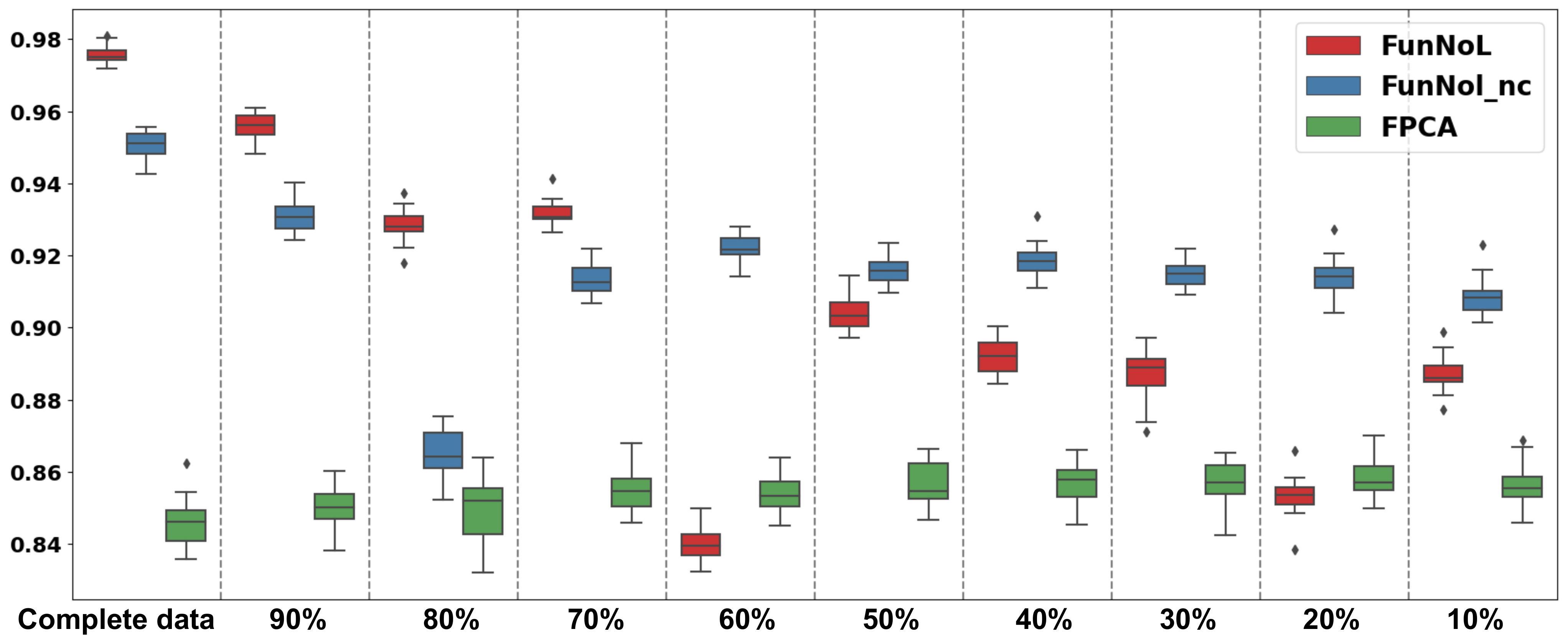}
    \caption{\label{fig:1D_classification_sparse} The simulations on \texttt{Starlight} data set. The y-axis represents the classification accuracy of test data whereas the x-axis corresponds to portions of available observations from individual functional data. The simulation is carried out 50 times for each sparsity setting. The distributions of prediction accuracy are represented by boxplots.}
\end{figure}
We conduct simulations on univariate functional data sets. The portion of correctly predicted labels given each setting is summarized in Figure~\ref{fig:1D_classification_sparse}. Figure 4 shows that the classification performance based on FunNoL representations does not suffer from the increasing portion of missing observations. Using the example of the \texttt{Starglight} data set, when the curves have more than 60\% data observed, adding a corruption step prevents overfitting the model and ensures that the representations have better generalization abilities hence are more useful in building prediction models. In this scenario, the FunNoL method with the corruption step has a larger prediction accuracy. On the other hand, adding manual corruptions to training data has a limit on improving the learned representations. When the curves have less than 60\% data observed, additional corruptions will overwhelm the estimation process, and the FunNoL method with no corruption step (\texttt{FunNOl\_nc}) has a higher prediction accuracy. Therefore, we recommend to use the FunNoL method with no corruption step in this scenario. 


\subsection{Simulation on multivariate functional data}
We will present simulation studies on multivariate functional data. The following simulations focused on the classification performance of the FunNoL method. 
\begin{figure}[H]
    \centering
    \subfloat{\includegraphics[width=.9\textwidth]{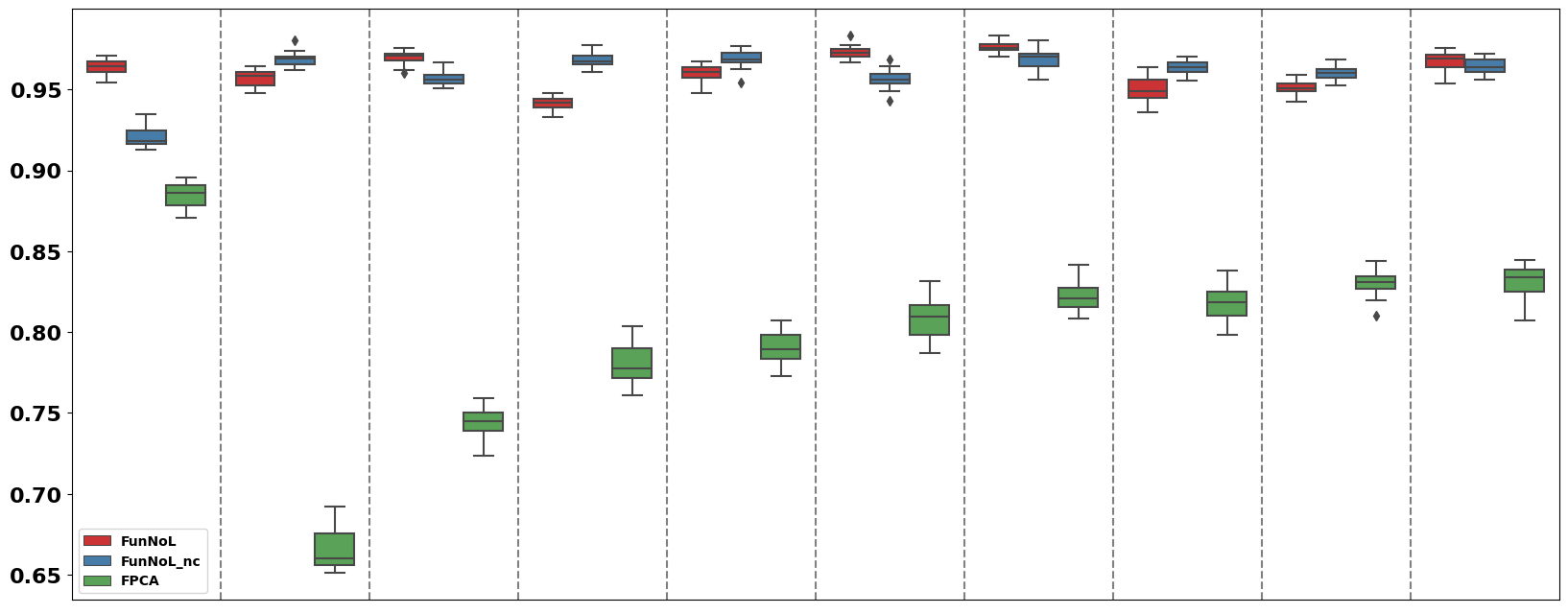}}\\
    \subfloat{\includegraphics[width=.9\textwidth]{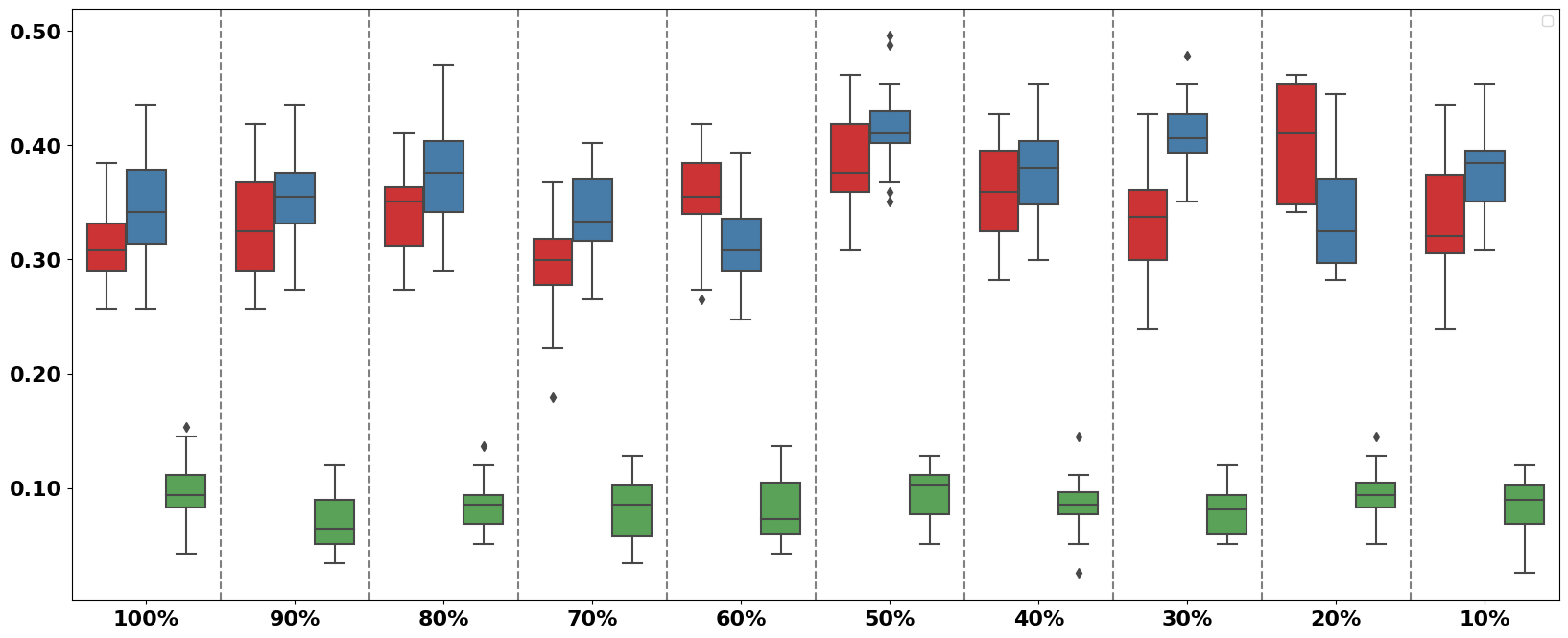}}
    \caption{\label{fig:3D_classificaiton} The top panel corresponds to the \texttt{gesture} data, and the bottom panel presents the classification of \texttt{cricket} data. Red boxes are the FunNoL model trained with corrupted observations, whereas blue boxes are the model trained with original observations. Green boxes represent the prediction accuracy by using FPC scores. Dashed lines separate different data settings where we sample a portion of training data to estimate the model. The leftmost panel corresponds to the case where all trajectories are used. The rightmost panel is the case where 10\% of data are sampled for doing estimation.}
\end{figure}
FunNoL is designed to address multivariate functional data, and we expect to see more significant improvement over FPCA as demonstrated in Figure~\ref{fig:3D_classificaiton}. FunNoL exceeds in predicting labels for both data sets. In addition, FunNoL performs equally well given increasing portion of missing observations. We can conclude that the proposed nonlinear mapping is more suitable for multivariate data where nonlinear structures become overwhelming for linear models. More simulation results are included in Section S8 of the supplementary document.

\section{Conclusion}\label{sec:Conclusion}
In this work, we propose a nonlinear learning method, FunNoL, for functional data. Its main purpose is to address the nonlinear structures in functional data and produce representations for future modeling tasks. It introduces flexible nonlinear mappings that provide better representations of functional data when linear projections are not sufficient. Also, the method can be directly applied to multivariate functional data. It achieves dimension reduction when the feature space of the mapping is a low-dimensional vector space. As a result, these vector representations reduce the efforts to build prediction models instead of directly using trajectories. The paper considers two modeling tasks, curve reconstruction and classification,  upon obtaining representations. We have applied FunNoL to several real data sets to examine its performances. In addition, we arranged different data corruption scenarios in simulation studies. The study results show that the FunNoL method is able to handle functional data with missing observations. The nonlinear mapping is also designed to be insensitive to any local disturbances in trajectories, hence the representations only retain the necessary information for subsequent prediction models. Conclusively, we have demonstrated the superiority of FunNoL as a nonlinear learning method for functional data. In this work, we focus on functional data defined on the time domain, which is well modeled by recurrent neural networks. Future extensions to this work can be made on functional data defined on multi-dimensional domain. For that matter, we can consider other neural networks to model multi-dimensional functional data.

\section*{Supplementary Materials}
A supplementary document includes the additional application and simulation results, proofs of the generalization bound, and some technical details. The application and simulation codes are published at \url{https://github.com/caojiguo/FunNoL}.

\bibliographystyle{Chicago}
\bibliography{FunNoL_JCGS}

\begin{thebibliography}{}

\bibitem[\protect\citeauthoryear{Bishop}{Bishop}{1995}]{trainingwithnoise}
Bishop, C.~M. (1995).
\newblock Training with noise is equivalent to tikhonov regularization.
\newblock {\em Neural Computation\/}~{\em 7\/}(1), 108--116.

\bibitem[\protect\citeauthoryear{Chen and Lei}{Chen and
  Lei}{2015}]{localizedFPCA}
Chen, K. and J.~Lei (2015).
\newblock Localized functional principal component analysis.
\newblock {\em Journal of the American Statistical Association\/}~{\em
  110\/}(511), 1266--1275.

\bibitem[\protect\citeauthoryear{Chiou, Chen, and Yang}{Chiou
  et~al.}{2014}]{chiouFPCA}
Chiou, J., Y.~Chen, and Y.~Yang (2014).
\newblock Multivariate functional principal component analysis: a normalization
  approach.
\newblock {\em Statistica Sinica\/}~{\em 24}, 1571--1597.

\bibitem[\protect\citeauthoryear{Dai and M\"{u}ller}{Dai and
  M\"{u}ller}{2018}]{rieFPCA}
Dai, X. and H.~M\"{u}ller (2018).
\newblock Principal component analysis for functional data on riemannian
  manifolds and spheres.
\newblock {\em The Annals of Statistics\/}~{\em 46}, 3334--3361.

\bibitem[\protect\citeauthoryear{Dau, Keogh, Kamgar, Yeh, Zhu, Gharghabi,
  Ratanamahatana, Yanping, Hu, Begum, Bagnall, Mueen, Batista, and
  Hexagon-ML}{Dau et~al.}{2018}]{UCRArchive2018}
Dau, H.~A., E.~Keogh, K.~Kamgar, C.-C.~M. Yeh, Y.~Zhu, S.~Gharghabi, C.~A.
  Ratanamahatana, Yanping, B.~Hu, N.~Begum, A.~Bagnall, A.~Mueen, G.~Batista,
  and Hexagon-ML (2018, October).
\newblock The ucr time series classification archive.

\bibitem[\protect\citeauthoryear{Dauxois, Pousse, and Romain}{Dauxois
  et~al.}{1982}]{fpca_asmp_3}
Dauxois, J., A.~Pousse, and Y.~Romain (1982).
\newblock Asymptotic theory for the principal component analysis of a vector
  random function: Some applications to statistical inference.
\newblock {\em Journal of Multivariate Analysis\/}~{\em 12}, 136--154.

\bibitem[\protect\citeauthoryear{Delaigle and Hall}{Delaigle and
  Hall}{2012}]{perfectclassification}
Delaigle, A. and P.~Hall (2012).
\newblock Achieving near perfect classification for functional data.
\newblock {\em Journal of the Royal Statistical Society: Series B (Statistical
  Methodology)\/}~{\em 74}, 267--286.

\bibitem[\protect\citeauthoryear{Hall and Hosseini-Nasab}{Hall and
  Hosseini-Nasab}{2005}]{fpca_asmp_1}
Hall, P. and M.~Hosseini-Nasab (2005).
\newblock On properties of functional principal components analysis.
\newblock {\em Journal of the Royal Statistical Society: Series B (Statistical
  Methodology)\/}~{\em 68}, 109--126.

\bibitem[\protect\citeauthoryear{Hall, Müller, and Wang}{Hall
  et~al.}{2006}]{fpca_asmp_2}
Hall, P., H.-G. Müller, and J.-L. Wang (2006).
\newblock Properties of principal component methods for functional and
  longitudinal data analysis.
\newblock {\em The Annals of Statistics\/}~{\em 34}, 1493--1517.

\bibitem[\protect\citeauthoryear{Happ and Greven}{Happ and
  Greven}{2018}]{mfpca_differentdomain}
Happ, C. and S.~Greven (2018).
\newblock Multivariate functional principal component analysis for data
  observed on different (dimensional) domains.
\newblock {\em Journal of the American Statistical Association\/}~{\em 113},
  649--659.

\bibitem[\protect\citeauthoryear{James, Hastie, and Sugar}{James
  et~al.}{2000}]{pcaforsparsefunctionaldata}
James, G., T.~Hastie, and C.~Sugar (2000).
\newblock Principal component models for sparse functional data.
\newblock {\em Biometrika\/}~{\em 87\/}(3), 587--602.

\bibitem[\protect\citeauthoryear{Kokoszak and Reimherr}{Kokoszak and
  Reimherr}{2017}]{kokoszkabook}
Kokoszak, P. and M.~Reimherr (2017).
\newblock {\em Introduction to Functional Data Analysis}.
\newblock Chapman and Hall/CRC.

\bibitem[\protect\citeauthoryear{Kong, Bondell, and Shen}{Kong
  et~al.}{2018}]{kongdehan_outlier}
Kong, D., H.~Bondell, and W.~Shen (2018).
\newblock Outlier detection and robust estimation in nonparametric regression.
\newblock In {\em Proceedings of the Twenty-First International Conference on
  Artificial Intelligence and Statistics}, Volume~84, pp.\  208--216.

\bibitem[\protect\citeauthoryear{Lin, Wang, and Cao}{Lin et~al.}{2015}]{iFPCA}
Lin, Z., L.~Wang, and J.~Cao (2015).
\newblock Interpretable functional principal component analysis.
\newblock {\em Biometrics\/}~{\em 72}, 846--854.

\bibitem[\protect\citeauthoryear{M\"{u}ller}{M\"{u}ller}{2005}]{Classification_fpca}
M\"{u}ller, H.-G. (2005).
\newblock Functional modelling and classification of longitudinal data.
\newblock {\em Scandinavian Journal of Statistics\/}~{\em 32}, 223--240.

\bibitem[\protect\citeauthoryear{M\"{u}ller and Stadtm\"{u}ller}{M\"{u}ller and
  Stadtm\"{u}ller}{2005}]{GFLM}
M\"{u}ller, H.-G. and U.~Stadtm\"{u}ller (2005).
\newblock Generalized functional linear models.
\newblock {\em The Annals of Statistics\/}~{\em 33}, 774--805.

\bibitem[\protect\citeauthoryear{Pascanu, Mikolov, and Bengio}{Pascanu
  et~al.}{2013}]{gradientclipping}
Pascanu, R., T.~Mikolov, and Y.~Bengio (2013).
\newblock On the difficulty of training recurrent neural networks.
\newblock In {\em Proceedings of the 30th International Conference on
  International Conference on Machine Learning}, Volume~28, pp.\  1310--1318.

\bibitem[\protect\citeauthoryear{Peng and M\"{u}ller}{Peng and
  M\"{u}ller}{2008}]{Clustering_FPCA}
Peng, J. and H.-G. M\"{u}ller (2008).
\newblock Distance-based clustering of sparsely observed stochastic processes,
  with applications to online auctions.
\newblock {\em The Annals of Applied Statistics\/}~{\em 2}, 1056--1077.

\bibitem[\protect\citeauthoryear{Ramsay and Silverman}{Ramsay and
  Silverman}{2005}]{fdabook}
Ramsay, J. and B.~Silverman (2005).
\newblock {\em Functional Data Analysis}.
\newblock New York: Springer.

\bibitem[\protect\citeauthoryear{Vieu and Ferraty}{Vieu and
  Ferraty}{2006}]{ferratybook}
Vieu, P. and F.~Ferraty (2006).
\newblock {\em Nonparametric Functional Data Analysis: Theory and Practice}.
\newblock New York: Springer.

\bibitem[\protect\citeauthoryear{Yao, M\"{u}ller, and Wang}{Yao
  et~al.}{2005a}]{sparseFPCA}
Yao, F., H.-G. M\"{u}ller, and J.-L. Wang (2005a).
\newblock Functional data analysis for sparse longitudinal data.
\newblock {\em Journal of the American Statistical Association\/}~{\em 100},
  577--590.

\bibitem[\protect\citeauthoryear{Yao, M\"{u}ller, and Wang}{Yao
  et~al.}{2005b}]{functional_linear}
Yao, F., H.-G. M\"{u}ller, and J.-L. Wang (2005b).
\newblock Functional linear regression analysis for longitudinal data.
\newblock {\em The Annals of Statistics\/}~{\em 33}, 2873--2903.

\bibitem[\protect\citeauthoryear{Zhou and Sang}{Zhou and
  Sang}{2021}]{centroidclassifier}
Zhou, Z. and P.~Sang (2021).
\newblock Continuum centroid classifier for functional data.
\newblock {\em Canadian Journal of Statistics\/}~{\em 50}.
\newblock \mbox{doi}: \url{https://doi.org/10.1002/cjs.11624}.

\end{thebibliography}
\end{document}